\documentclass[aps,pre,showpacs,twocolumn,preprintnumbers,amsmath,amsfonts,bm]{revtex4-1}
\usepackage{graphicx}
\usepackage{subfigure}
\usepackage{color}
\usepackage{amssymb}

\def\w{2.5in}

\newcommand{\T}{\intercal}
\begin{document}

\title{Memory and Information Processing in Recurrent Neural Networks}

\author{Alireza~Goudarzi$^{1}$,~Sarah~Marzen$^{2}$,~Peter~Banda$^{3}$,~Guy~Feldman$^{4}$, Matthew~R.~Lakin$^{1}$,~Christof Teuscher$^{5}$ and ~Darko~Stefanovic$^{1}$}

\affiliation{
$^1$University of New Mexico, Albuquerque, NM 87131\\
$^2$University of California Berkeley, Berkeley, CA 94720\\
$^3$University of Luxembourg,~Belvaux,~Luxembourg L-4367\\
$^4$Purdue University, West Lafayette, IN 47907\\
$^5$Portland State University, Portland, OR 97206
}

\date{\today}

\begin{abstract}
Recurrent neural networks (RNN) are simple  dynamical systems whose computational power has been attributed to their short-term memory. Short-term memory of RNNs has been previously studied analytically  only for the case of orthogonal networks, and only under annealed approximation, and uncorrelated input.  Here for the first time, we present an exact solution to the memory capacity and the task-solving performance as a function of the structure of a given network instance, enabling direct determination of the function--structure relation in RNNs. We calculate the memory capacity for arbitrary networks with exponentially correlated input and further related it to the performance of the system on  signal processing tasks in a supervised learning setup. We compute the expected error and the worst-case error bound as a function of the spectra of the network and the correlation structure of its inputs and outputs. Our results give an explanation for learning and generalization of task solving using short-term memory, which is crucial for building alternative computer architectures using physical phenomena based on the short-term memory principle.  
\end{abstract}


\maketitle
Excitable dynamical systems, or reservoirs, store a short-term memory of a driving input signal in their instantaneous  state~\citep{ganguli2008}. This memory can  produce a desired output in a linear readout layer, which can be trained efficiently using ordinary linear  regression or  gradient descent. This paradigm, called~\protect{\em reservoir computing} (RC), was originally proposed as a simplified model of information processing in the prefrontal cortex~\protect\citep{dominey1995}. It  was later generalized to explain computation in cortical microcircuits~\protect\citep{Maass:2002p1444} and to facilitate training in recurrent neural networks~\protect\citep{jaeger2004}. A central feature of RC is the lack of fine tuning of the underlying dynamical system: any random structure that guarantees a stable dynamics   gives rise to short-term memory~\protect\citep{Maass:2002p1444,jaeger2004}. Analogous behavior has also been observed in selective response in random neural populations~\protect\citep{Hansel2012}. Furthermore, fixed underlying structure in RC makes it suitable for implementing computation using  spatially distributed physical phenomena\citep{goudarzi2012,goudarzi2013,0957-4484-24-38-384004,burger2015,Nakajima20140437,Vandoorne:2014qe,PhysRevE.91.020801,7180580}. Such approaches can give us a way to store and process information more efficiently than with von Neumann architecture~\citep{crutchfield2010}.



  
Short-term memory in   neural networks has been studied for uncorrelated input $u(t)$ under annealed approximation, i.e., connectivity is resampled independently at each time step~\citep{white2004}. That study considered only linear orthogonal networks, where the columns of the connectivity matrix are pairwise orthogonal and the node transfer functions are linear. A memory function $m(\tau)$ was defined to measure  the ability of the system to reconstruct input from  $\tau$ time steps ago, i.e., $u(t-\tau)$, from the present system state $x(t)$. It was shown that the total memory capacity cannot exceed the $N$ degrees of freedom in the system. For networks with saturating nonlinearity, the memory scales with $\sqrt N$~\citep{ganguli2008}; however, by fine-tuning the nonlinearity one can achieve near-linear scaling of memory capacity~\citep{Toyoizumi:2012kq}. In nonlinear networks, it is very difficult to analyze  the complete memory function and even harder to relate it to the performance on computational tasks, as is  evident from many works in this area with hard-to-reconcile conclusions (see Ref.~\citep{lukosevicius2012}).
\vspace{2mm}
{\bf\em Model}---
Consider a discrete-time network of $N$ nodes. The network weight matrix $\boldsymbol\Omega$ is  $N\times N$  with spectral radius $\lambda<1$. A time-dependent scalar input signal $u_t$ is fed to the network using the input weight vector $\boldsymbol\omega$.  The  evolution of the network state $x_t$ and the output $y_t$ is governed by
\begin{align}
x_{t+1} &= \boldsymbol\Omega x_t + \boldsymbol\omega u_t, \text{ and}\\
y_{t+1} &=\boldsymbol\Psi x_{t+1},
\label{eq:timeevolution}
\end{align}
where $\boldsymbol\Psi =  \left({\bf XX^\T}\right)^{-1}  {\bf X} {\bf \widehat{Y}^\T}$ is an $N$-dimensional column vector calculated for a desired output $\widehat{y}_t$.
Here, each  column of ${\bf X}$ is the state of the network at time $x_t$ and each column of $\widehat{\bf Y}^\T$ is the corresponding desired output at each time step. In practice it is sometimes necessary to use Tikhonov regularization to calculate the readout weights, i.e., $\boldsymbol\Psi =  \left({\bf XX^\T} + \gamma^2 {\bf I}\right)^{-1}  {\bf X} {\bf \widehat{Y}^\T}$, where $\gamma$ is a regularization factor that needs to be adjusted depending on $\boldsymbol\Omega,\boldsymbol\omega$, and $u_t$ \cite{lukosevicius2012}.\\

Calculating $\boldsymbol\Psi$ for a given problem requires the following input-output-dependent evaluations (Appendix~\ref{app:computing_xx_xy}):
\protect\begin{align}
  {\bf X}{\bf X}^\T
 &= \sum_{i,j=0}^{\infty} \boldsymbol\Omega^i\boldsymbol\omega R_{uu}(i-j) \boldsymbol\omega^{\T}(\boldsymbol\Omega^\T)^{j} , \text{ and }\label{eq:genericXX} \\
  {\bf X}{\bf Y}^\T
 &= \sum_{i=0}^{\infty} \boldsymbol\Omega^i \boldsymbol\omega R_{u\widehat{y}}(i),
 \label{eq:basicsums}
\end{align}
where $R_{uu}(i-j) = \langle u_{t} u_{t-(i-j)} \rangle$ and $R_{u\widehat{y}}(i-j) = \langle u_{t} \widehat{y}_{t-(i-j)} \rangle$ are  the autocorrelation of the input and the cross-correlation of the input and target output. This may also be expressed  more generally in terms of the power spectrum of the input and the target: 
\begin{align} 
{\bf X}{\bf X}^\T &= \frac{1}{2T} \int_{-T}^{T}\boldsymbol\Omega_+^{-1}\boldsymbol\omega S_{uu}(f) \boldsymbol\omega^{\T}\boldsymbol\Omega_-^{-1} df,\\
{\bf X}{\bf Y}^\T &= \frac{1}{2T}\int_{-T}^{T} \boldsymbol\Omega_+^{-1} \boldsymbol\omega S_{u\widehat{y}}(f) e^{if \tau} df.
\end{align}
where $\boldsymbol\Omega_{+}=\left({\bf I}-e^{if}\boldsymbol\Omega \right)$ and $\boldsymbol\Omega_{-}=\left({\bf I}-e^{-if}\boldsymbol\Omega \right)$, $S_{uu}(f)$ is the power spectral density of the input, and $S_{u\widehat{y}}(f)$ is the cross-spectral density of the input and the target output.

The performance can be evaluated by the mean-squared-error (MSE) as follows:
\protect\begin{align}
\langle E^2\rangle &= \left\langle \left(\widehat{y}(t)-y(t)\right)^2 \right\rangle = \widehat{\bf Y}\widehat{\bf Y}^\T - \widehat{\bf Y}{\bf Y}\\
 &= \widehat{\bf Y}\widehat{\bf Y}^\T -  \widehat{\bf Y}{\bf X}^\T  \left( {\bf XX^\T}\right)^{-1}    {\bf X}\widehat{\bf Y}^\T.
\end{align}

The  MSE gives us a distribution-independent upper bound on the instantaneous-squared-error through the application of Markov inequality:
\protect\begin{align}
P\left[ \left(\widehat{y}(t)-y(t)\right)^2 \ge a\right] \le \frac{\langle E^2\rangle}{a}.
\end{align}
The existence of a worst-case bound is  important for engineering applications of RC. 

\vspace{2mm}
{\bf\em Memory and task solving}--- 
The memory function of the system is defined as~\citep{white2004}
 \begin{align}
 m(\tau) &=  ({\bf Y}{\bf X}^\T)({\bf X}{\bf X}^\T){^{-1}}({\bf X}{\bf Y}^\T),
 \end{align}
 where ${\bf Y}$ is the input with lag $\tau$, $u_{t-\tau}$.

 The exact evaluation of this function has previously proved elusive for arbitrary $\boldsymbol\Omega$ and $\boldsymbol\omega$. Here we provide a solution using eigendecomposition of $\boldsymbol\Omega$ and the power spectral density of the input signal. The solution may also be described directly in terms of the autocorrelation of the input, as  in Appendix~\ref{app:computing_mcac}.

Let $\boldsymbol\Omega = {\bf U} \text{diag}({\bf d}) {\bf U}^{-1}$ and $\bar{\boldsymbol\omega} = {\bf U}^{-1}\boldsymbol\omega$
so that
\begin{align}
\left({\bf I}-e^{if}\boldsymbol\Omega\right)^{-1} &= {\bf  U} \text{diag}(\frac{1}{1-e^{if}{\bf d}}) {\bf  U}^{-1}.
\end{align}
The memory function is reduced to
\begin{align}
m(\tau) &= \frac{1}{2T} \bar{\boldsymbol\omega}^{\top} \left(A \circ C^{-1} \right) \bar{\boldsymbol\omega},
\end{align}
where the matrix $C$ is given by
\begin{align}
C &= \int_{-T}^{T} [\frac{\bar{\boldsymbol\omega}}{1-e^{if} {\bf d}}]\otimes[\frac{\bar{\boldsymbol\omega}}{1-e^{-if} {\bf d}}] S_{uu}(f) df,
\end{align}
and the matrix $A$ is given by
\begin{align}
A &= [\int_{-T}^{T}\frac{S_{u\widehat{y}}(f)e^{if\tau}}{1-e^{if} {\bf  d}}df] \otimes [\int_{-T}^{T}\frac{S_{u\widehat{y}}(f)e^{if\tau}}{1-e^{if} {\bf d}}df].
\end{align}

The total memory is then given by
\begin{align}
\sum_{\tau=0}^{\infty} m(\tau) 
&= \frac{1}{2T} \bar{\boldsymbol\omega}^{\top} \left(B \circ C^{-1}\right) \bar{\boldsymbol\omega}
\end{align}
where
\begin{align}
B &= \int_{-T}^{T}\int_{-T}^{T} \frac{df df'}{1-e^{i(f+f')}} [\frac{S_{u\widehat{y}}(f)}{1-e^{if}{\bf  d}}] \otimes [\frac{S_{u\widehat{y}}(f')}{1-e^{if'}{\bf d}}].
\end{align}
 \begin{figure}[ht]
\def\w{250px}
\def\h{123px}
\centering
\subfigure[]{
\includegraphics[width=\w,height=\h]{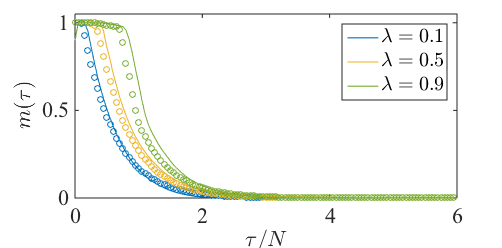}
}\\
\subfigure[]{
\includegraphics[width=\w,height=\h]{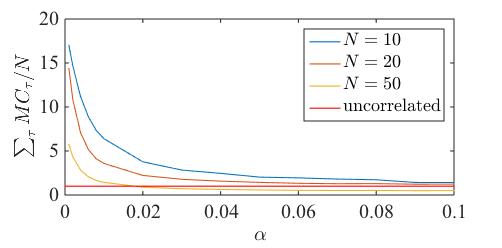}
}
\caption{(a) Agreement of analytical and empirical memory function for different $\lambda$. (b) Scaling of memory capacity with increasing structure in the input.}
\label{fig:correlated_input}
\end{figure}
\noindent

\begin{figure}[h]
\centering
\def\w{250px}
\def\h{123px}
\subfigure[]{
\includegraphics[width=\w,height=\h]{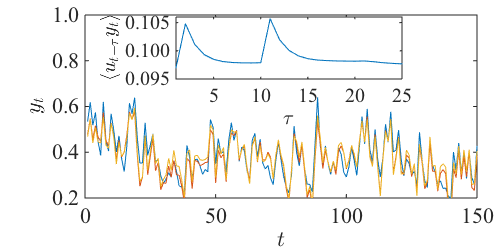}
\label{fig:narma10}}\\
\subfigure[]{
\includegraphics[width=\w,height=\h]{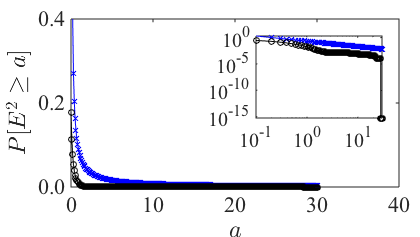}
\label{fig:narma10bound}
}
\caption{Target output and system output generated with analytical weights and trained weights for the NARMA10 task (a), the worst-case bounds using the Markov inequality, with the same plot on a log-log scale in the inset (b).}
\label{fig:narma10corr}
\end{figure}

\begin{figure}[h]
\centering
\def\w{250px}
\def\h{123px}
\subfigure[]{
\includegraphics[width=\w,height=\h]{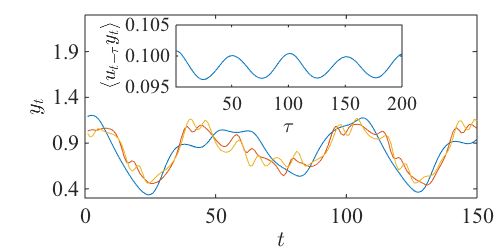}
\label{fig:mg}}\\
\subfigure[]{
\includegraphics[width=\w,height=\h]{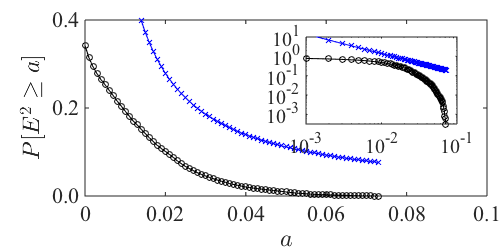}
\label{fig:mgbound}
}
\caption{ Target output and system output generated with analytical weights and trained weights for the Mackey-Glass 10 step ahead prediction task (c), the worst-case bounds using the Markov inequality, with the same plot on a log-log scale in the inset (d).}
\label{fig:mgcorr}
\end{figure}

We validate our formula by comparing analytical and  empirical evaluation of the memory curve. The input is assumed to be a sequence of length $T$ with autocorrelation function $R(\tau) = e^{-\alpha \tau}$ (Appendix~\ref{appsec:setup}).
FIG~\ref{fig:correlated_input}(a) shows the result of the single-instance calculation of the analytical and the empirical  memory curves for different $\lambda$ (Appendix~\ref{appsec:setup}). As expected, the analytical and empirical results agree. 

We also study the memory capacity for different levels of structure in the input signal. Here, we use simple ring topologies with $\lambda=0.9$ and vary the decay exponent $\alpha$, FIG~\ref{fig:correlated_input}(b). For a fixed system size, decreasing $\alpha$ exponentially increases the correlation in the input, which increases the memory capacity.

Next, we use our method to calculate the optimal output layer, the expected average error, and bounds on worst-case error for a common NARMA10 benchmark task:
\begin{equation}
y_t=\alpha y_{t-1}+\beta y_{t-1} \sum_{i=1}^{n}y_{t-i}+\gamma u_{t-n}u_{t-1}+\delta,
\end{equation}
where $n=10$, $\alpha=0.3, \beta=0.05, \gamma=1.5, \delta=0.1$. The input $u_t$ is
drawn from a uniform distribution over the interval $[0,0.5]$. Here the evaluation of ${\bf X}{\bf X}^\T$ follows the same calculation as for the memory capacity for the uniform distribution. For ${\bf X}{\bf Y}^\T$ we must estimate the cross-correlation of $y_t$ and $u_t$ and substitute it into   Equation~\ref{eq:basicsums}.
 FIG~\ref{fig:narma10corr}(a) shows the output of a network of $N=20$ nodes and $\lambda=0.9$ with analytically calculated optimal output layer. The output agrees  with the correct output of the NARMA10 system. The cross-correlation of the system used for the calculation is shown in the inset. FIG~\ref{fig:narma10corr}(b) shows the worst-case error bound for this system and the empirical errors generated from the system output, showing that the bound we derived is tight.

\begin{figure}[h]
\def\w{250px}
\def\h{123px}
\centering
\subfigure[]{
\includegraphics[width=\w,height=\h]{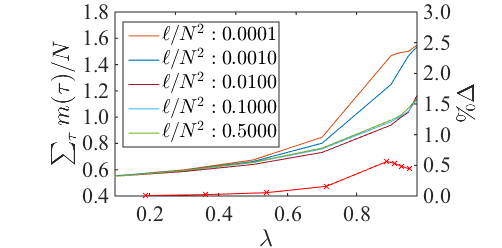}
\label{fig:N100_mclambda}}\\
\subfigure[]{
\includegraphics[width=\w,height=\h]{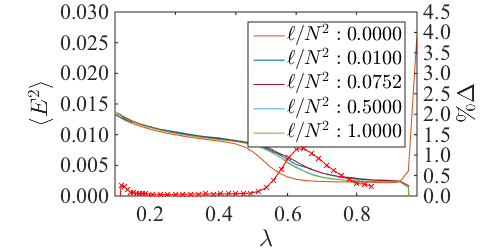}
\label{fig:narmalambda}
}\\
\subfigure[]{
\includegraphics[width=\w,height=\h]{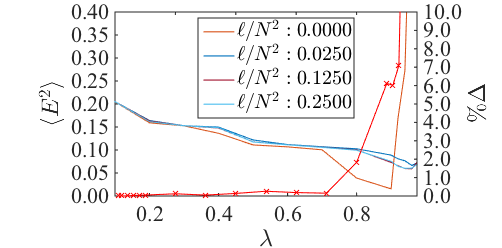}
\label{fig:MGN200_lambda}
}
\caption{Total memory capacity (a), NARMA10 error (b), and Mackey-Glass prediction error (c) as a function of $\lambda$ and increasing network irregularity. The best results in all cases are for completely regular networks.}
\label{fig:toplogyresult}
\end{figure}

Finally, we show how the framework can be used in a prediction scenario, namely the prediction of the Mackey-Glass system. The Mackey-Glass system \citep{Mackey15071977} was first proposed as a model for feedback systems that may show different dynamical regimes. The system is a one-dimensional delayed feedback differential equation and manifests a wide range of dynamics,  from fixed points to strange attractors with varying divergence rates (Lyapunov exponent). This system has been used as a benchmark task for chaotic signal prediction  and generation~\citep{jaeger2004}. It  is defined as:
\begin{equation}
\frac{dx(t)}{dt}=\beta \frac{x(t-\tau)}{1+x(t-\tau)^n}-\gamma x(t),
\label{eq:mgeq}
\end{equation}
where $\beta=2, \gamma=1, n=9.7451, \tau=2$ ensure the chaoticity of the dynamics~\citep{jaeger2004}.


FIG~\ref{fig:narma10corr}(c) shows the prediction result for 10 time steps ahead and the inset shows the  autocorrelation  at different lags. The autocorrelation is characterized by a long correlation length evident from non-zero correlation values for large $\tau$. This long memory is a hallmark of chaotic systems. We use this information to evaluate   Equation~\ref{eq:genericXX} and   Equation~\ref{eq:basicsums}, where for predicting $\tau$ steps ahead we have ${\bf X}{\bf Y}^\T= \sum_{i=\tau}^{\infty} \boldsymbol\Omega^i \boldsymbol\omega R_{u\widehat{y}}(i).$

\vspace{2mm}
{\bf\em The effect of network structure}---
The  effect of randomness and sparsity of reservoir connectivity has been a subject of debate~\citep{lukosevicius2012}.
To study the effect of network structure on memory and performance, we systematically explore the range between  sparse deterministic uniform networks and random graphs. We start from a simple ring topology with identical weights and induce noise to $\ell$ random links by sampling the normal distribution $\mathcal{N}(0,1)$. We then re-evaluate the memory and task solving performance keeping the weight matrix fixed.
We evaluate system performance on a memory task with exponentially correlated inputs, the nonlinear autoregressive NARMA10 task, and the Mackey-Glass chaotic time series prediction. We  systematically explore the effects of $\lambda$, $N$, and $\ell$ on the performance of the system.


   FIG~\ref{fig:toplogyresult}(a)
       shows the resulting total memory capacity normalized by $N$ as a function of increasing randomness $\frac{\ell}{N^2}$ for different spectral radii $\lambda$. The expected theoretical total memory capacity for an uncorrelated signal is $\sum_\tau m(\tau)/N=1$. Here the system exploits the structure of the input signal to store longer input sequences, i.e., $\sum_\tau m(\tau)/N>1$. This effect has been studied previously under annealed approximation and in a compressive sensing setup ~\citep{NIPS2010_3980}. However, here we see that even without the sparse input assumption and  $L_1$ optimization in the output (a computationally expensive optimization used in compressive sensing) the network can achieve  capacity greater than its degrees of freedom $N$. FIG~\ref{fig:toplogyresult}(b) and (c) show the error in the NARMA10 and the Mackey-Glass prediction tasks. Here,  best performance is achieved for a regular architecture. A slight randomness significantly increases error at first, but additional irregularity will decrease it. This can be observed for the NARMA10 task at $\lambda=0.6$ and for the Mackey-Glass prediction task at $\lambda=0.9$. 
\vspace{2mm}
{\bf\em Discussion}---
Although memory capacity of RNNs has been studied before, it is learning and generalization ability in a task solving setup is not discussed. Our derivation allows us to relate the memory capacity to task solving performance for arbitrary RNNs and reason about their generalization. In empirical experiments with systems presented here, the training and testing are done with finite input sequences that are sampled independently for each experiment, so the statistics of the training and testing inputs vary according to a Gaussian distribution around their true values and one expects these estimates to approach their true values with increasing sample size. Hence, the mean-squared-error $\langle E^2 \rangle$, which is  linear in the input and output statistics, is also distributed as a Gaussian for repeated experiments. By the law of large numbers, the difference between testing and training mean-squared-error tends to zero in the limit.  This  explains the ability of the system to generalize its computation from training to test samples.  
 

\vspace{2mm}
{\bf\em Conclusion}---
The computational power of reservoir computing networks has been attributed to their memory capacity. While their memory properties have been studied under annealed approximation, thus far no direct mathematical connection to their signal processing performance had been made. We developed a mathematical framework to exactly calculate the memory capacity of RC systems and extended the framework to study their expected and worst-case errors on a given task in a supervised learning setup. Our result confirms previous studies that the upper bound for memory capacity for uncorrelated inputs is $N$. We further show that the memory capacity monotonically increases with correlation in the input. Intuitively, the output exploits the redundant structure of the inputs to retrieve longer sequences.  Moreover, we generalize our derivation to task solving performance. Our derivation help us reason about the memory and performance of arbitrary systems  directly in terms of their structure.  We  showed that networks with regular structure have a higher memory capacity but are very sensitive  to  slight changes in the structure, while irregular networks are robust to variation in their structure. 

{\em Acknowledgments.}
This work was partly supported by NSF grants \#1028238 and \#1028120, \#1518861, and \#1525553.

\appendix

\section{Computing the optimal readout weights using autocorrelation}
\label{app:computing_xx_xy}

The recurrent neural network (RNN) model that we study in this paper is an echo state network (ESN) with linear activation function. This system consist of an input driven recurrent network of size $N$, and a linear readout layer trained to calculate a desired function of the input. Let $u(t)$ and $\boldsymbol\omega$ indicate a one-dimensional  input at time $t$ and an input weight vector respectively. Let $\boldsymbol\Omega$ be a $N\times N$ recurrent weight matrix, ${\bf x}(t)$ be an $N$-dimensional network state at time $t$, and $\boldsymbol\Psi$ be the readout weight vector. The dynamics of the network and output is described by: 
\begin{align} 
{\bf x}(t+1) &= \boldsymbol\Omega {\bf x}(t) + \boldsymbol\omega  u(t) \\
y(t+1) &= \boldsymbol\Psi {\bf x}(t+1), 
\end{align} 
where the readout weights are given by \cite{white2004}: 
\begin{align} 
\boldsymbol\Psi &= ({\bf X}{\bf X}^\top)^{-1}{\bf X}{\bf Y}^\top.
\end{align}

The value of the optimal readout weights depend on the covariance and cross-covariance components $({\bf X}{\bf X}^\top)$ and ${\bf X}{\bf Y}^\top$. Here we show that these can be computed exactly for any arbitrary system given by $\boldsymbol\Omega$ and $\boldsymbol\omega$ and autocorrelation of the input $R_{uu}(\tau)$ and cross-correlation of input and output $R_{u\widehat{y}}$.

We begin by noting that the explicit expression for the system state is given by: 
\begin{align} 
{\bf x}(t+1) &= \sum_{k=0}^{t}\boldsymbol\Omega^{t-k}\boldsymbol\omega u_{k}.
\end{align}
Calculating $\Psi$ for a given problem requires the following input-output-dependent evaluations:
\protect\begin{align}
  {\bf X}{\bf X}^\T
    &= \left\langle   \sum_{i,j=0}^{\infty} \boldsymbol\Omega^{i} \boldsymbol\omega u_{t-i}  u^\T_{t-j}\boldsymbol\omega^\T {\boldsymbol\Omega^\T}^{j} \right\rangle \nonumber \\
    &= \sum_{i,j=0}^{\infty} \boldsymbol\Omega^i\boldsymbol\omega R_{uu}(i-j) \boldsymbol\omega^{\T}(\boldsymbol\Omega^\T)^{j} , \text{ and }\label{eq:genericXXmain} \\
  {\bf X}{\bf Y}^\T
    &= \left\langle   \sum_{i=0}^{\infty} \boldsymbol\Omega^{i} \boldsymbol\omega u_{t-i}  \widehat{y}^\T_{t}\right\rangle \nonumber \\
    &= \sum_{i=0}^{\infty} \boldsymbol\Omega^i \boldsymbol\omega R_{u\widehat{y}}(i).
 \label{eq:basicsumsmain}
\end{align}

\section{Computing the total memory using autocorrelation}
\label{app:computing_mcac}

Here we compute the memory function and the total memory of the recurrent neural network described in Appendix~\ref{app:computing_xx_xy} for exponentially correlated input where $R_{uu}(\tau) = e^{-\alpha \tau}$.
The total memory of the system is given by the following summation over the memory function \cite{white2004}: 

\begin{align}
\sum_\tau m(\tau) = \text{Tr}( ({\bf X}{\bf X}^\T)^{-1} \sum_{\tau=0}^\infty  ({\bf X}{\bf Y}^\T)_\tau ({\bf Y}{\bf X}^\T)_\tau).
\end{align}

 where ${\bf Y}$ is the input with lag $\tau$, $u_{t-\tau}$.
 
Computing ${\bf X}{\bf X}^\T$ requires the evaluation of:
\begin{align}
{\bf X}{\bf X}^\T &= \sum_{i,j=0}^{\infty} \boldsymbol\Omega^i\boldsymbol\omega R_{uu}(i-j) \boldsymbol\omega^{\T}(\boldsymbol\Omega^\T)^{j}.
\end{align}
This assumes an even correlation function, i.e., $R_{uu}(i-j)=R_{uu}(j-i)$. For numerical computation it is more convenient to perform the calculation as follows:
\begin{align}
{\bf X}{\bf X}^\T &= {\bf X}{\bf X}^\T_{i\geq j} + {\bf X}{\bf X}^\T_{i\leq j} - {\bf X}{\bf X}^\T_{i=j} \nonumber \\
 &= {\bf X}{\bf X}^\T_{i\geq j} + \left({\bf X}{\bf X}^\T_{i\geq j}\right)^\T - {\bf X}{\bf X}^\T_{i=j},
\end{align}
where ${\bf X}{\bf X}^\T_{i\geq j}$ is a partial sum of ${\bf X}{\bf X}^\T$ satisfying $i\geq j$, $ {\bf X}{\bf X}^\T_{i\leq j}= \left({\bf X}{\bf X}^\T_{i\geq j}\right)^\T$ is a partial sum of ${\bf X}{\bf X}^\T$ satisfying $i\leq j$, and ${\bf X}{\bf X}^\T_{i=j}$ is a partial sum of ${\bf X}{\bf X}^\T$ satisfying $i=j$, which is double counted and must be subtracted. We can substitute $\tau=|i-j|$ and evaluate ${\bf X}{\bf X}^\T_{i\geq j}$ and  ${\bf X}{\bf X}^\T_{i=j}$ as follows:
\begin{align}
{\bf X}{\bf X}^\T_{i\geq j} &= \sum_{i,\tau=0}^\infty \boldsymbol\Omega^i \boldsymbol\omega  R_{uu}(\tau) \boldsymbol\omega^\T (\boldsymbol\Omega^\T){^{i+\tau}}, \text{ and} \\
{\bf X}{\bf X}^\T_{i=j} &=    \sum_{i=0}^\infty \boldsymbol\Omega^i \boldsymbol\omega  R_{uu}(0) \boldsymbol\omega^\T (\boldsymbol\Omega^\T){^{i}}.
\end{align}

\begin{align}
{\bf X}{\bf X}^\T_{i\geq j} &   = \sum_{i,\tau=0}^\infty \boldsymbol\Omega^i \boldsymbol\omega  R_{uu}(\tau) \boldsymbol\omega^\T (\boldsymbol\Omega^\T){^{i+\tau}}\\
   &= \sum_{i,\tau=0}^\infty \boldsymbol\Omega^i \boldsymbol\omega  e^{-\alpha \tau} \boldsymbol\omega^\T (\boldsymbol\Omega^\T){^{i+\tau}}\\
      &=  \sum_{i=0}^\infty \boldsymbol\Omega^i \boldsymbol\omega   \boldsymbol\omega^\T (\boldsymbol\Omega^\T){^{i}} \sum_{\tau=0}^\infty (e^{-\alpha}\boldsymbol\Omega^\T )^\tau\\
      &= {\bf U } {\boldsymbol\Lambda } \circ ({\bf  I}^\circ  - {\bf d}{\bf d}^\T)^{-1^\circ} {\bf U}^\T
       ({\bf I} - e^{-\alpha} \boldsymbol\Omega^\T)^{-1},\\
       & \nonumber\\
       {\bf X}{\bf X}^\T_{i=j} &=    \sum_{i=0}^\infty \boldsymbol\Omega^i \boldsymbol\omega  R_{uu}(0) \boldsymbol\omega^\T (\boldsymbol\Omega^\T){^{i}}\\
       &= \sum_{i=0}^\infty \boldsymbol\Omega^i \boldsymbol\omega   \boldsymbol\omega^\T (\boldsymbol\Omega^\T){^{i}}\\
       &=   {\bf U } {\boldsymbol\Lambda } \circ ({\bf  I}^\circ  - {\bf d}{\bf d}^\T)^{-1^\circ} {\bf U}^\T.
\end{align}

Here ${\bf I}^\circ$ is the identity of the Hadamard product denoted by $\circ$, and $^{-1^\circ}$ is a matrix inverse with respect to the Hadamard product. Here the trick is that $\bar{\boldsymbol\omega} = {\bf U}^{-1} \boldsymbol\omega$ takes the input to the basis of the connection matrix $\boldsymbol\Omega$ allowing the dynamics to be described by the powers of the eigenvalues of $\boldsymbol\Omega$, i.e., ${\bf D}$. Since ${\bf D}$ is symmetric we can use the matrix identity ${\bf D}{ \boldsymbol\Lambda} {\bf D}=\boldsymbol\Lambda \circ {\bf d}{\bf d}^\T$, where ${\bf d}$ is the main diagonal of ${\bf D}$. Summing over the powers of ${\bf D}$ gives us $\sum_{i=0}^\infty \boldsymbol\Omega^i \boldsymbol\omega   \boldsymbol\omega^\T (\boldsymbol\Omega^\T){^{i}}= {\bf U } {\boldsymbol\Lambda } \circ ({\bf  I}^\circ  - {\bf d}{\bf d}^\T)^{-1^\circ} {\bf U}^\T$.

The covariance of the network states and the expected output is given by:
\begin{align}
  {\bf X}{\bf Y}^\T_\tau
    &=    \sum_i \boldsymbol\Omega^{i} \boldsymbol\omega R(| i - \tau|)
    = \sum_i \boldsymbol\Omega^{i} \boldsymbol\omega e^{-\alpha| i - \tau|}.
  \end{align}
For $\alpha\to\infty$, the signal becomes  i.i.d.  and the calculations simplify as follows \citep{Goudarzi2014176}:
 \begin{align}
 {\bf X}{\bf X}^\T  & = \langle u^2 \rangle  {\bf U } {\boldsymbol\Lambda } \circ ({\bf  I}^\circ  - {\bf d}{\bf d}^\T)^{-1^\circ} {\bf U}^\T,\\
 {\bf X}{\bf Y}^\T &=   \boldsymbol\Omega^{\tau} \boldsymbol\omega \langle u^2 \rangle.
\end{align}

The total memory capacity can be calculated by summing over $m(k)$:
\begin{align}
\sum_\tau m(\tau) = \text{Tr}( ({\bf X}{\bf X}^\T)^{-1} \sum_{\tau=0}^\infty  ({\bf X}{\bf Y}^\T)_\tau ({\bf Y}{\bf X}^\T)_\tau).
\end{align}

\section{Experimental Setup for Memory Task}
\label{appsec:setup}
For our experiment with memory capacity of network under exponentially correlated input we used the following setup. 
We generated $T=2,000,000$ long sample inputs with autocorrelation function $R_{uu}(\tau) = e^{-\alpha\tau}$. To generate exponentially correlated input we draw $T$ samples ${u}_i$ from a uniform distribution over the interval $[0,1]$. The samples are
   passed through a low-pass filter with a smoothing factor $\alpha$. We normalize and center $u_t$ so that $\langle u(t) \rangle_t=0$ and $\langle u(t)^2 \rangle_t=1$. The resulting normalized samples $u(t)$ have exponential autocorrelation with decay exponent $\alpha$, i.e., $R_{uu}(\tau)=e^{-\alpha \tau}$. To validate our calculations, we use a network of $N=20$ nodes in a ring topology and identical weights. The spectral radius $\lambda=0.9$. The input weights $\boldsymbol\omega$ are created by sampling the binomial distribution and multiplying with $0.1$. The scale of the input weights does not affect  the memory and the performance in linear systems and therefore we adopt this convention for generating $\boldsymbol\omega$ throughout the paper.  We also assumed $\alpha=0.05$, the number of samples $T=30,000$,  washout period of $5,000$ steps, and regularization factor $\gamma^2=10^{-9}$.

\section{Experimental Setup for Topological Study}
\label{appsec:topology}
A long standing question in recurrent neural network is how its structure effect its memory and task solving performance. Our derivation lets us compute optimal readout layer for arbitrary network. Here we describe the calculations we performed to  examine the effect of structure of the network on its memory and task solving performance. To this end, we use networks of size $N=100$, $\alpha=0.01$, and $\gamma=10^{-13}$ and we systematically study the randomness and spectral radius. We start from a uniform weight ring topology  and incrementally add randomness from $\ell=0$ to $\ell= \frac{N^2}{2}$. The results for each value of $\ell$ and $\lambda$ are averaged over $50$ instances. This averaging is necessary even for $\ell=0$ because the input weights are randomly generated and although their scaling does not affect the result their exact values do~\citep{ganguli2008}.

\section{Computing the optimal readout weights using power spectrum}
\label{appsec:analyticspsdgeneric}

The calculations in Appendix~\ref{app:computing_xx_xy} for optimal layer of a recurrent network may be described in a more generally in terms of power spectrum of the input signal. Here we assume the setup  in Appendix~\ref{app:computing_xx_xy} and derive an expressions for optimal readout layer using its the power spectrum of the input and output. 

We start by the standard calculation of ${\bf X}{\bf X}^\T$ and ${\bf X}{\bf Y}^\T$:
\begin{align}
  {\bf X}{\bf X}^\T
    &= \left\langle   \sum_{i,j=0}^{\infty} \boldsymbol\Omega^{i} \boldsymbol\omega u_{t-i}  u^\T_{t-j}\boldsymbol\omega^\T {\boldsymbol\Omega^\T}^{j} \right\rangle \nonumber \\
    &= \sum_{i,j=0}^{\infty} \boldsymbol\Omega^i\boldsymbol\omega R_{uu}(i-j) \boldsymbol\omega^{\T}(\boldsymbol\Omega^\T)^{j} , \text{ and }\label{eq:genericXXmain} \\
  {\bf X}{\bf Y}^\T
    &= \left\langle   \sum_{i=0}^{\infty} \boldsymbol\Omega^{i} \boldsymbol\omega u_{t-i}  \widehat{y}^\T_{t}\right\rangle \nonumber \\
    &= \sum_{i=0}^{\infty} \boldsymbol\Omega^i \boldsymbol\omega R_{u\widehat{y}}(i).
 \label{eq:basicsumsmain}
\end{align}

We replace
\begin{align}
R_{uu}(t) &= \frac{1}{2T}\int_{-T}^{T} S_{uu}(f) e^{{if t}} df
\end{align}
and
\begin{align}
R_{u\widehat{y}}(t) &= \frac{1}{2T}\int_{-T}^{T} S_{u\widehat{y}}(f) e^{if t} df
\end{align}
which gives
\begin{widetext}
\begin{align}
  {\bf X}{\bf X}^\T
    &= \frac{1}{2T}\int_{-T}^{T}\sum_{t,t'=0}^{\infty} \boldsymbol\Omega^t\boldsymbol\omega S_{uu}(f) e^{if (t-t')} \boldsymbol\omega^{\T}(\boldsymbol\Omega^\T)^{t'} \\
    &= \frac{1}{2T}\int_{-T}^{T}\left(\sum_{t=0}^{\infty} (e^{if}\boldsymbol\Omega)^t \right)\boldsymbol\omega S_{uu}(f) \boldsymbol\omega^{\T}\left(\sum_{t'=0}^{\infty} (e^{-if}\boldsymbol\Omega^\T)^{t'}\right) df \\
    &= \frac{1}{2T} \int_{-T}^{T}\left(I-e^{if}\boldsymbol\Omega \right)^{-1}\boldsymbol\omega S_{uu}(f) \boldsymbol\omega^{\T}\left(I-e^{-if}\boldsymbol\Omega^\T\right)^{-1} df
\end{align}
\end{widetext}
and

\begin{align}
({\bf X}{\bf Y}^\T)_{\tau}
    &= \frac{1}{2T}\int_{-T}^{T}\sum_{t=0}^{\infty} \boldsymbol\Omega^t \boldsymbol\omega S_{u\widehat{y}}(f) e^{if (t+\tau)} df \\
    &= \frac{1}{2T}\int_{-T}^{T} \left(\sum_{t=0}^{\infty} (e^{if}\boldsymbol\Omega)^t\right) \boldsymbol\omega S_{u\widehat{y}}(\omega) e^{if\tau} df \\
    &= \frac{1}{2T}\int_{-T}^{T} \left(I-e^{if}\boldsymbol\Omega\right)^{-1} \boldsymbol\omega S_{u\widehat{y}}(f) e^{if \tau} df.
\end{align}

\section{Memory capacity expressed in terms of power spectrum}
\label{appsec:analyticspsdmc}

Here we use the derivation in Appendix~\ref{appsec:analyticspsdgeneric}  and compute the memory function and the total memory of the system. Let $\boldsymbol\Omega = \boldsymbol U \text{diag}({\bf d}) \boldsymbol U^{-1}$ and $\bar{\boldsymbol\omega} = {\bf U}^{-1}\boldsymbol\omega$
so that
\begin{align}
\left(I-e^{if}\boldsymbol\Omega\right)^{-1} &= \boldsymbol U \text{diag}(\frac{1}{1-e^{if}{\bf d}}) \boldsymbol U^{-1}.
\end{align}

We find that
\begin{align}
m(\tau) &= \frac{1}{2T} \bar{\boldsymbol\omega}^{\top} \left(A \circ C^{-1} \right). \bar{\boldsymbol\omega}
\end{align}
The matrix $C$ is given by
\begin{align}
C &= \int_{-T}^{T} [\frac{\bar{\boldsymbol\omega}}{1-e^{if} {\bf d}}]\otimes[\frac{\bar{\boldsymbol\omega}}{1-e^{-if} {\bf d}}] S_{uu}(f) df,
\end{align}
and the matrix $A$ is given by: 
\begin{align}
A &= [\int_{-T}^{T}\frac{S_{u\widehat{y}}(f)e^{if\tau}}{1-e^{if} {\bf  d}}df] \otimes [\int_{-T}^{T}\frac{S_{u\widehat{y}}(f)e^{if\tau}}{1-e^{if} {\bf d}}df].
\end{align}
The total memory is then given by: 
\begin{align}
\frac{1}{N}\sum_{\tau=0}^{\infty} m(\tau) 
&= \frac{1}{2TN} \bar{\boldsymbol\omega}^{\top} \left(B \circ C^{-1}\right) \bar{\boldsymbol\omega}
\end{align}
where
\begin{align}
B &= \int_{-T}^{T}\int_{-T}^{T} \frac{df df'}{1-e^{i(f+f')}} [\frac{S_{u\widehat{y}}(f)}{1-e^{if}{\bf  d}}] \otimes [\frac{S_{u\widehat{y}}(f')}{1-e^{if'}{\bf d}}].
\end{align}

\end{document}